\documentclass[sigconf]{acmart}
\usepackage{multirow}
\usepackage{cases}
\usepackage{array}
\usepackage{url}
\usepackage{makecell}
\usepackage{tabularx}
\usepackage{graphicx}
\usepackage{enumitem}
\usepackage{marvosym}
\AtBeginDocument{%
  }

\copyrightyear{2023}
\acmYear{2023}
\setcopyright{acmcopyright}
\acmConference[WSDM '23] {Proceedings of the Sixteenth ACM International Conference on Web Search and Data Mining}{February 27-March 3, 2023}{Singapore, Singapore.}
\acmBooktitle{Proceedings of the Sixteenth ACM International Conference on Web Search and Data Mining (WSDM '23), February 27-March 3, 2023, Singapore, Singapore}
\acmPrice{15.00}
\acmISBN{978-1-4503-9407-9/23/02}
\acmDOI{3539597.3570479}

\begin{document}
\fancyhead{}

\title{Long-Document Cross-Lingual Summarization}

\author{Shaohui Zheng$^{1*}$, Zhixu Li$^{2* (\textrm{\Letter})}$, Jiaan Wang$^{1}$, Jianfeng Qu$^{1(\textrm{\Letter})}$ \\
An Liu$^{1}$, Lei Zhao$^{1}$, and Zhigang Chen$^{3}$}

\makeatletter
\def\authornotetext#1{
 \g@addto@macro\@authornotes{%
 \stepcounter{footnote}\footnotetext{#1}}%
}
\makeatother

\authornotetext{The first two authors made equal contributions to this work.}

\affiliation{%
  \institution{$^{1}$ School of Computer Science and Technology, Soochow University, Suzhou, China}
  \country{}
}
\affiliation{%
  \institution{$^{2}$ Shanghai Key Laboratory of Data Science, School of Computer Science, Fudan University, Shanghai, China}
  \country{}
}
\affiliation{%
  \institution{$^{3}$ Jilin Kexun Information Technology Co., Ltd., Jilin, China}
  \country{}
}

\email{{shzheng3, jawang1}@stu.suda.edu.cn,zhixuli@fudan.edu.cn}
\email{{jfqu, anliu, zhaol}@suda.edu.cn, zgchen@iflytek.com}

\begin{abstract}
Cross-Lingual Summarization (CLS) aims at generating summaries in one language for the given documents in another language. CLS has attracted wide research attention due to its practical significance in the multi-lingual world. Though great contributions have been made, existing CLS works typically focus on short documents, such as news articles, short dialogues and guides.
Different from these short texts, long documents such as academic articles and business reports usually discuss complicated subjects and consist of thousands of words, making them non-trivial to process and summarize.
To promote CLS research on long documents, we construct Perseus, the first long-document CLS dataset which collects about 94K Chinese scientific documents paired with English summaries. The average length of documents in Perseus is more than two thousand tokens.
As a preliminary study on long-document CLS, we build and evaluate various CLS baselines, including pipeline and end-to-end methods.
Experimental results on Perseus show the superiority of the end-to-end baseline, outperforming the strong pipeline models equipped with sophisticated machine translation systems.
Furthermore, to provide a deeper understanding, we manually analyze the model outputs and discuss specific challenges faced by current approaches.
We hope that our work could benchmark long-document CLS and benefit future studies.
\end{abstract}



\maketitle

\section{Introduction}

\begin{sloppypar}
Given documents in a source language, Cross-Lingual Summarization (CLS) aims to generate the corresponding summaries in a different target language.
Under the background of globalization, CLS could help people obtain key information from documents in their unfamiliar languages, making information acquisition more efficient.
Consequently, this task becomes more important and has attracted wide research attention~\cite{Wang2022ASO}.

Nevertheless, current CLS works generally focus on short texts. For example, Zhu et al.~\cite{zhu-etal-2019-ncls} propose two CLS datasets, En2ZhSum and Zh2EnSum, and Bai et al.~\citep{2021Cross} propose En2DeSum. These three widely-used CLS datasets are all collected from English and Chinese news reports with the scales of 371K, 1.7M, and 438K, respectively.
The average lengths of source documents in En2ZhSum and En2DeSum are 755.0 and 31.0 words (in English), respectively, while the counterpart in Zh2EnSum is 103.7 characters (in Chinese).
Besides, Ladhak et al.~\cite{ladhak-etal-2020-wikilingua} construct WikiLingua with an average of 45K CLS samples per cross-lingual direction\footnote{We use ``direction'' to denote the summarization direction from the source to the target languages, e.g., English (documents) $\Rightarrow$ Chinese (summaries). Some CLS datasets contain more than one direction, thus we show the average samples per direction.}, and the average length of their source documents is 391 words.
Perez-Beltrachini and Lapata~\cite{perez-beltrachini-lapata-2021-models} construct XWikis which involves 214K documents per direction and the average length of their source documents is 945 words.
Recently, Wang et al.~\cite{2022ClidSum} propose a dialogue-oriented CLS dataset named XSAMSum which contains 16K dialogue documents whose average length is 83.9 words.
Different from these short texts, long documents usually provide detailed discussions of multiple topics and involve more than thousands of words.
Building a long-document CLS system has practical significance since it can save a lot of reading time for people who are not familiar with the source language.
However, this task is still under-explored due to the lack of corresponding datasets.

\begin{figure*}[t]

\centerline{\includegraphics[width=0.93\textwidth]{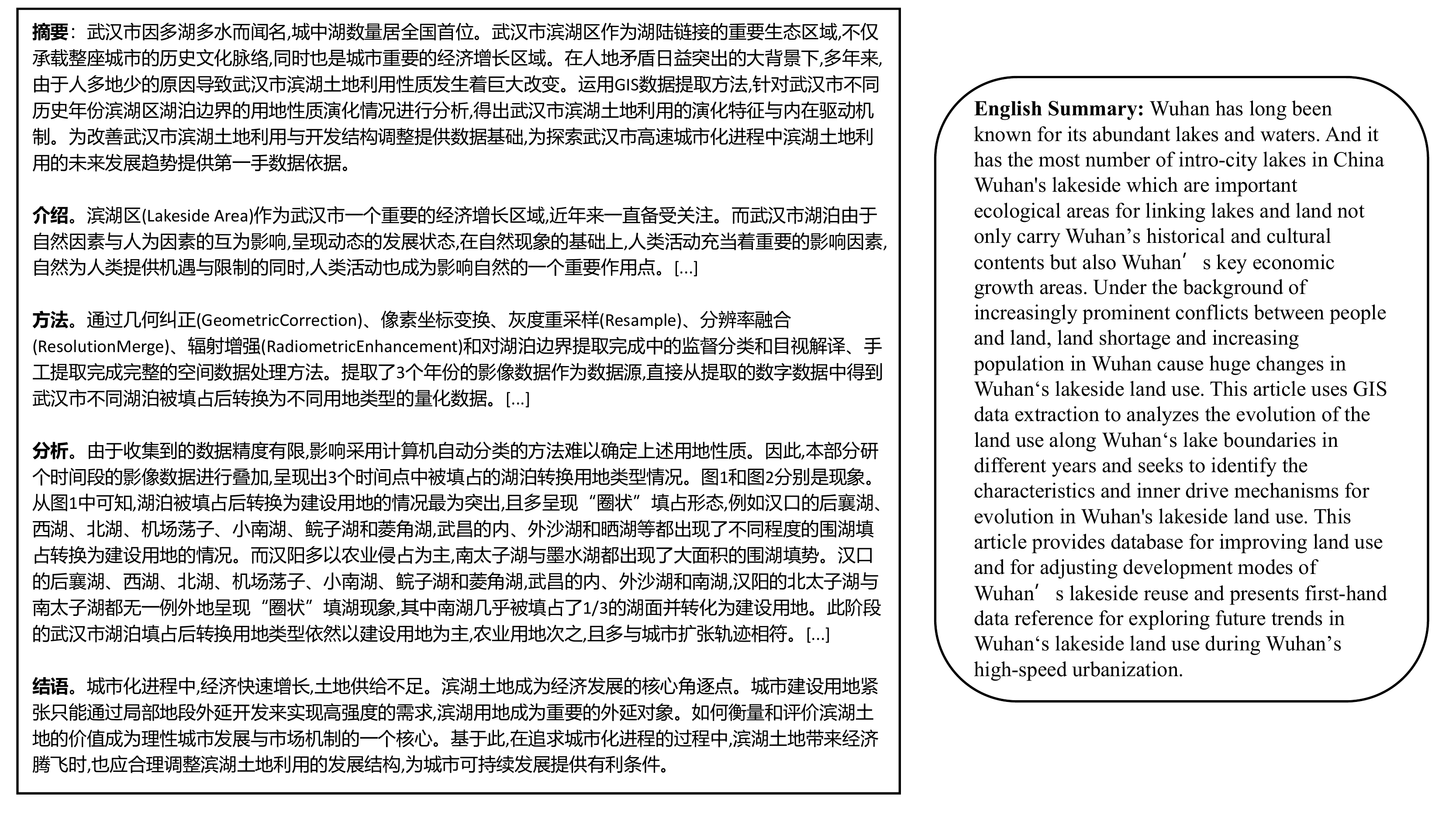}}
\caption{An example of Chinese long document and the target English summary.}
\label{fig:intro_case}
\end{figure*}

In this paper, we construct {\bf Perseus}\footnote{The dataset has been released at \url{https://github.com/LearnItBoy/Perseus}} (Scientific Pa\textbf{pers} Onlin\textbf{e} for Cross-Ling\textbf{u}al \textbf{S}ummarization), the first long-document CLS dataset which contains about 94K Chinese scientific papers paired with English summaries (an example is shown in Figure~\ref{fig:intro_case}).
Our dataset covers four disciplines in total, including engineering applications, natural science, agricultural science and medical science.
The average length of its source documents is 2872.9 Chinese characters, which is significantly larger than those of previous datasets~\cite{zhu-etal-2019-ncls,2021Cross,ladhak-etal-2020-wikilingua,perez-beltrachini-lapata-2021-models,2022ClidSum}.
To evaluate the generalization of long document CLS models, we also provide an out-of-domain test set that contains 500 Chinese documents together with their English summaries pairs in the sports domain.
In view of pre-trained language models, most of them can only handle hundreds of tokens (e.g., 512-token limitation in BERT-style NLU models and 1,024-token limitation in BART-style NLG models) due to the quadratic memory and computational consumption, thus it is a great challenge to model the long-distance dependencies within long documents.
Besides, long documents usually need long summaries to convey their core ideas, leading to more difficulty in generating comprehensive and accurate summaries.

Existing CLS methods generally follow three paradigms, i.e., translate-then-summarize~\cite{Leuski2003CrosslingualCE,wan-2011-using,yao-etal-2015-phrase,Zhang2016AbstractiveCS,ouyang-etal-2019-robust}, summarize-then-translate~\cite{Orasan2008EvaluationOA,wan-etal-2010-cross}, and end-to-end~\cite{zhu-etal-2019-ncls,cao-etal-2020-jointly,xu-etal-2020-mixed,liang-etal-2022-variational,feng-etal-2022-msamsum,2022ClidSum}. Specifically, the translate-then-summarize methods first translate the source documents into the target language and then summarize the translated documents.
In contrast, summarize-then-translate methods first summarize the source-language documents and further translate the summaries into the target language. End-to-end methods directly generate summaries in the target language from the given documents in the source language.
Among them, the translate-then-summarize paradigm is not suitable for our scenario due to the high costs caused by machine translation in long documents (it needs to translate the whole documents instead of brief summaries).
Thus, we build and evaluate various summarize-then-translate and end-to-end methods based on Perseus. 
In addition, we divide the summarize-then-translate methods into \textit{extract-then-translate} methods and \textit{abstract-then-translate} methods depending on whether the summarization methods is extractive (directly select sentences from documents as summaries) or abstractive (use sequence-to-sequence models to generate summaries).
Experimental results on Perseus show that the end-to-end baseline performs best in terms of all metrics (including automatic and human evaluation metrics), demonstrating its superiority in generating logical, informative and concise summaries.

Moreover, to provide a deeper understanding of long-document CLS, we manually analyze the model outputs and summarize the main challenges brought by this new task, including (\romannumeral1) missing information, (\romannumeral2) redundancy, (\romannumeral3) wrong references and (\romannumeral4) semantically unclear generation. We hope that our work could prompt the CLS research on long documents and inspire future studies.

Our main contributions are concluded as follows:
\begin{itemize}[leftmargin=*,topsep=0pt]
    \item To the best of our knowledge, we are the first CLS work on long documents. We construct the first long-document CLS dataset named Perseus, containing 94K Chinese long documents and the corresponding English summaries. An out-of-domain test set is also provided to evaluate the model's generalization.
    \item Based on the Perseus dataset, we build and evaluate various baselines from different paradigms and manually analyze the model outputs to provide deeper analyses.
    \item We conduct thorough analyses of this task based on Perseus and discuss the promising directions for future work.
\end{itemize}
\end{sloppypar}

\section{Related Work}

\subsection{Cross-Lingual Summarization}
\begin{sloppypar}
Cross-Lingual Summarization (CLS) has received a lot of research attention these years, and many valuable datasets and methods have been proposed one after another. Based on existing monolingual summarization datasets, Zhu et al.~\cite{zhu-etal-2019-ncls} design a round-trip translation strategy with a machine translation service to construct the first large-scale CLS datasets, i.e., En2ZhSum and Zh2EnSum. Later, Bai et al.~\cite{2021Cross} construct En2DeSum in the same way.
The target summaries of these three datasets are all machine translated from other languages.
Recently, some CLS datasets are constructed through manually translating the summaries of existing monolingual summarization datasets to different target languages, e.g., ClidSum~\cite{2022ClidSum} employs professional translators to translate the summaries of two English monolingual dialogue summarization to German and Chinese.
GOAL~\cite{wang2022goal} manually translates the summaries of its collected English monolingual sports summarization data to Chinese.
In another way, researchers also attempt to collect CLS data from multi-lingual online resources.
Global Voice~\cite{nguyen-daume-iii-2019-global}, WikiLingua~\cite{ladhak-etal-2020-wikilingua} and XWikis~\cite{perez-beltrachini-lapata-2021-models} crawl multi-lingual document-summary pairs from Global Voice, WikiHow, and Wikipedia websites, respectively.

Early CLS methods typically focus on pipeline paradigms, i.e., translate-then-summarize~\cite{Leuski2003CrosslingualCE,wan-2011-using,yao-etal-2015-phrase,Zhang2016AbstractiveCS,ouyang-etal-2019-robust} and summarize-then-translate~\cite{Orasan2008EvaluationOA,wan-etal-2010-cross}. The former first translates the source documents to the target language and then summarizes the translated documents. And the latter first generates the summaries of the source documents and further translates them to the target language.
Though straightforward, these pipeline methods suffer from severe propagation errors.
Recently, many efforts~\cite{zhu-etal-2019-ncls,cao-etal-2020-jointly,xu-etal-2020-mixed,liang-etal-2022-variational,feng-etal-2022-msamsum,2022ClidSum,Wang2022ASO} are given to prompt end-to-end CLS methods.
\end{sloppypar}

\subsection{Long-Document Summarization}
Long-document summarization aims to generate a summary from a given long document containing thousands of tokens.
Cohan et al.~\cite{cohan2018discourse} present two large-scale long-document summarization datasets, arXiv and PubMed, whose data are collected from the corresponding scientific paper websites, arXiv.org and PubMed.com. In addition, there are many other long-document summarization datasets collected from various resources, including patents~\cite{2019BIGPATENT}, government reports~\cite{2021Efficient} and sports games~\cite{Wang2021SportsSum20GH,wang2022knowledge}.
It is difficult for traditional summarization methods to perform long documents due to their limited capability of modeling long-distance dependencies.
Transformer-based pre-trained models can only take no more than 1024 tokens as input because the memory and compute consumption of self-attention scales quadratically with input length.
To enlarge the maximum acceptable input length of pre-trained models, longformer encoder-decoder~\cite{2020Longformer} uses a local windowed attention mechanism with a task-guided global attention to make the consumption scales linearly as input length. Besides, there are many other sparse attention mechanisms~\cite{2020Efficient,2021Efficient} that can efficiently process long sequences.

\section{Dataset}
In this section, we first introduce the construction process of our dataset (\S~\ref{subsec:data_collection}). To verify the generalization of long-document CLS models, we also present an out-of-domain test set (\S~\ref{subsec:ood_test}). Finally, we give data statistics to provide deeper analyses (\S~\ref{subsec:data_analyses}).

\subsection{Data Construction}
\label{subsec:data_collection}
\begin{sloppypar}
\noindent \textbf{Resource}. We crawl long-document CLS data from \textit{Sciencepaper Online}\footnote{\url{https://www.paper.edu.cn/}}, which records large amounts of Chinese scientific papers.
We choose these Chinese scientific papers due to: (1) Scientific papers usually reach thousands of words and could be regarded as long documents. There are many monolingual long-document summarization datasets leveraging scientific papers as their research object~\cite{cohan2018discourse}. (2) Many Chinese academic journals require researchers to write abstracts in both Chinese and English~\cite{Wang2022ASO}. Thus, the English summaries and the corresponding Chinese paper contents could naturally form CLS samples.
We crawl 652 journals from 2000 to 2021, covering four subjects, i.e., engineering applications, natural science, agricultural science, and medical science. As a result, 418.7k papers (in PDF format) are collected.

\vspace{0.5ex}
\noindent \textbf{Pre-Processing.} To extract the Chinese documents, Chinese summaries and English summaries from the science papers in PDF format, we employ the following data pre-processing:

We first utilize PyMuPDF toolkit\footnote{\url{https://github.com/pymupdf/PyMuPDF}} to translate PDF papers to plain texts, and then extract Chinese summaries and English summaries via the rule-based methods.
The texts extracted by the toolkit contain a lot of noise such as journal information that is irrelevant to the articles. Thus, we design a lot of regular expression templates to filter this noise. Following ArXiv dataset~\cite{cohan2018discourse}, we remove the reference information and replace reference endnotes with a special token, we also use another special token to replace the math formulations in the documents.
We also find that the toolkit works well with Chinese characters, but suffers from a problem where there is no space between English words when parsing English contents.
To address this problem, we transform the first pages of papers into pictures and employ tesseract-OCR\footnote{\url{https://github.com/tesseract-ocr/tesseract}} to extract their English summaries.

After the pre-processing, we remove the documents whose lengths are less than 1,000 or over 8,000, and discard papers that do not contain either Chinese or English summaries. Finally, we obtain 94K triples of $\langle$Chinese document, Chinese summary, English summary$\rangle$, which constitute our Perseus.
We split our dataset into 82K/6K/6K w.r.t training/validation/test set.

\vspace{0.5ex}
\noindent \textbf{Quality.} We randomly select 200 samples from the test set and employ three graduate students as evaluators to check these samples. The evaluators are asked to give one point for a sample if there is no noise, zero otherwise.
Finally, we receive 176 points on average, achieving a noise-free rate of 88\%.

\end{sloppypar}

\subsection{Out-of-Domain Test}
\label{subsec:ood_test}
\begin{sloppypar}
We also provide an out-of-domain test set (OOD set) to evaluate the model's generalization.
To this end, we choose K-SportsSum~\cite{wang2022knowledge}, a Chinese long-document summarization dataset in the sports domain. 
To adapt this dataset to the CLS task, we manually translate the summaries in its test set (with 500 samples) from Chinese to English.
In detail, we have three graduate students who are all native Chinese speakers with fluent English to translate these summaries. After manually translating, a data expert will check the translating results to make sure the translations are qualified.
In this way, the original Chinese documents paired with the translated English summaries could form the out-of-domain test set.
\end{sloppypar}

\begin{table*}[]
    \centering

    \resizebox{0.9\textwidth}{!}
    {
    \begin{tabular}{lcccccccc}
    \toprule[1pt]
        \multicolumn{1}{c|}{\textbf{Dataset}} & \textbf{Domain} & \makecell{\textbf{Doc}\\\textbf{Num.}} & \makecell{\textbf{Src}\\\textbf{Lang.}} &  \makecell{\textbf{Tgt}\\\textbf{Lang.}} & \makecell{\textbf{Doc.}\\\textbf{Length}} & \makecell{\textbf{Src Summ.}\\\textbf{Length}} & \makecell{\textbf{Tgt Summ.}\\\textbf{Length}} & \makecell{\textbf{Comp.}\\\textbf{Ratio}}\\
    \bottomrule[1pt]
    \multicolumn{8}{c}{Previous CLS Datasets} \\
    \toprule[1pt]
    \multicolumn{1}{l|}{Eh2ZnSum~\cite{zhu-etal-2019-ncls}} & News Report & 371K  & En & Zh
    & 755.0 & 55.2 & 96.0 & 13.7\\
    \multicolumn{1}{l|}{Zn2EhSum~\cite{zhu-etal-2019-ncls}} & News Report & 1.7M & Zh & En
    & 103.7 & 17.9 & 13.7 & 5.8\\
    \multicolumn{1}{l|}{En2DeSum~\cite{2021Cross}}          & News Report & 438K  & En & De
    & 31.0  & 8.5  & 7.5  & 3.6\\
    \multicolumn{1}{l|}{XSAMSum~\cite{2022ClidSum}}   & Dialogue & 16K   & En & De/Zh
    & 83.9  & 20.3 & 19.9/33.0 & 4.1\\
    \multicolumn{1}{l|}{WikiLingua~\cite{ladhak-etal-2020-wikilingua}}   & How-to Guide & 46K   & Multi & Multi
    & 391.0  & / & 39.0 & /\\
    \multicolumn{1}{l|}{XWikis~\cite{perez-beltrachini-lapata-2021-models}}   &  Encyclopedia Article & 214K   & Multi & Multi
    & 945.0  & / & 77.0 & /\\
    \bottomrule[1pt]
    \multicolumn{8}{c}{Perseus} \\
    \toprule[1pt]
    \multicolumn{1}{l|}{Train}    & Scientific Paper  & 82K  & Zh & En & 2871.2 & 201.2 & 124.1 & 14.3\\
    \multicolumn{1}{l|}{Validate} & Scientific Paper  & 6K   & Zh & En & 2880.5 & 199.7 & 122.8 & 14.4\\
    \multicolumn{1}{l|}{Test (in-domain)}     & Scientific Paper  & 6K   & Zh & En & 2883.4 & 202.3 & 124.7 & 14.3\\
    \multicolumn{1}{l|}{Test (out-of-domain)}     & Sports Game & 0.5K & Zh & En & 3970.9 & 612.1 & 456.3 & 6.5 \\
    \bottomrule[1pt]
    \end{tabular}
    }
    
    \setlength{\belowcaptionskip}{10pt}
    \caption{Data statistics of Perseus and previous CLS datasets. \textit{Doc Num.} is the number of samples in each dataset. \textit{Src Lang.} and \textit{Tgt Lang.} denote the source language and the target language, respectively (En: English, Zh: Chinese, De: German). \textit{Doc. Length}, \textit{Src Summ. Length} and \textit{Tgt Summ. Length} indicate the average lengths of source documents, source-language summaries and target-language summaries, respectively. The lengths are counted in word level for English and character level for Chinese. \textit{Comp. Ratio} represents compression ratio. \textbf{WikiLingua} and \textbf{XWikis} have multiple source-target language pairs and their average lengths are averaged over CLS samples in all cross-lingual directions.}
    \label{tab:dataset_information}
\end{table*}

\subsection{Statistics}
\label{subsec:data_analyses}

Table~\ref{tab:dataset_information} shows the data statistics of our Perseus as well as previous CLS datasets. The average lengths of the documents in Perseus are 2871.2, 2880.5 and 2883.4 w.r.t training, validation and test sets, respectively, longer than all previous CLS datasets.
We also calculate compression ratio for these datasets, which is the result of the average length of source-langauge documents divided by the average length of source-language summaries. It reveals how much the summary refines the content of the document. Our compression ratio of our dataset is 14.3/14.4/14.3 (training/validation/test), larger than all previous CLS datasets, which also means the documents in our dataset contain more redundant information and the distribution of their key information is sparser.
%
The out-of-domain test set is in the sports domain, and we do not limit their lengths to the same level as those of Perseus, thus, we can evaluate the generalization of CLS models trained on Perseus.

\section{Baselines}
In this section, we first formally definite the long-document CLS task (\S~\ref{subsec:4.1}), then we introduce the details of various baselines including extract-then-translate (\S~\ref{subsec: ext-trans}), abstract-then-translate (\S~\ref{subsec:4.3}) and end-to-end (\S~\ref{subsec:4.4}) baselines.

\subsection{Task Definition}
\label{subsec:4.1}
Long-document cross-lingual summarization aims to generate a brief summary $S=\{t_1, t_2, ..., t_{|S|}\}$ in a target language given a long document $D=\{s_1, s_2, ..., s_{|D|}\}$ in a different source language, where $t_i$ denotes the $i$-th token and $s_j$ denotes the $j$-th sentence.

\subsection{Extract-then-Translate}
\label{subsec: ext-trans}
\begin{sloppypar}
Extract-then-Translate (Ext-Trans) is a pipeline paradigm that directly extracts Chinese sentences from a document and translates these sentences into English to obtain the target summaries.
A key component lies in Ext-Trans is extractor which directly selects sentences from original documents as their summaries. In this way, the summaries involve few grammatical errors, but lose flexibility. We adopt the following four extractors in the Ext-Trans methods:
\begin{itemize}[leftmargin=*,topsep=0pt]
\item \textbf{Longest} is a heuristic way to directly select the longest $k$ sentences from each document as its summary.
\item \textbf{TextRank}~\cite{mihalcea-tarau-2004-textrank} is an unsupervised sentence-level ranking algorithm based on undirected graph. 
\item \textbf{PacSum}~\cite{zheng-lapata-2019-sentence} could be regarded as an upgraded version of TextRank. It uses the position information in the graph network to judge the pointing relationship between sentences, so as to convert the traditional undirected graph into a directed graph and improve the model performance of selecting key sentences.
\item \textbf{SummaRu.}~\cite{Nallapati2017SummaRuNNerAR} is a supervised RNN-based extracting method.
\end{itemize}

Given the source-language documents, we select the top-5 sentences based on each extractor to form their source-language summaries.
Next, we adopt the following machine translation (MT) methods (including sophisticated MT service and open source MT model) to translate the summaries to the target language:
\begin{itemize}[leftmargin=*,topsep=0pt]
\item \textbf{Baidu Translation\footnote{\url{https://fanyi-api.baidu.com/}}} is a sophisticated MT service.
\item \textbf{OPUS-MT}~\cite{tiedemann-thottingal-2020-opus} releases many MT models with the architecture of transformer. we utilize the pre-trained OPUS-MT-zh-en model\footnote{\url{https://huggingface.co/Helsinki-NLP/opus-mt-zh-en}} to translate summaries from Chinese into English.
\end{itemize}
\end{sloppypar}

\subsection{Abstract-then-Translate}
\label{subsec:4.3}
\begin{sloppypar}
Abstract-then-Translate (Abs-Trans) first employs a sequence-to-sequence (seq2seq) model to generate the summaries of the given source-language documents and then translates the summaries from the source to the target language.
Abstractive methods can flexibly generate summaries conditioned on the key information of the document.
A key component lies in Abs-Trans is abstractor with the architecture of seq2seq. Specifically, we adopt the following two seq2seq models as abstractors, respectively:
\begin{itemize}[leftmargin=*,topsep=0pt]
\item \textbf{PGN}~\cite{see-etal-2017-get} is a LSTM-based seq2seq model, introducing copy mechanism and coverage mechanism to alleviate the problems of out-of-vocabulary and repeated generation. PGN can take long documents as inputs but cannot efficiently model the long-distance dependencies due to its LSTM-based architecture.
\item \textbf{LED} (Longformer-Encoder-Decoder)~\cite{2020Longformer} is a seq2seq model with sparse attention mechanism. The weights of LED is initialized by BART~\cite{lewis-etal-2020-bart} (a pre-trained transformer-based seq2seq model). LED is suitable for processing long documents due to its sparse attention (based on sliding window attention).

\end{itemize}
The above abstractors are trained with monolingual document-summary pairs (in Chinese). Next, we adopt the same MT methods as Ext-Trans to translate the summaries from Chinese to English.
\end{sloppypar}

\subsection{End-to-End}
\label{subsec:4.4}
The end-to-end method directly generates a target-language summary given a source-language document in a seq2seq manner.
To build the end-to-end baseline, we modify mBART-50~\cite{2020Multilingual}, a state-of-the-art multilingual generative model which is pre-trained on a large-scale multi-lingual corpus involving 50 languages, to support the inputs of long documents.
In detail, we replace the dense self-attention in vanilla mBART-50 with LED-style sparse self-attention. It could also be regarded as a multi-lingual version of LED (denoted as \textbf{mLED}).

\section{Experiments}
\subsection{Implementation Details}
\label{imp_details}
\begin{sloppypar}

The pre-trained models in our experiments are provided by the Huggingface Transformers Library\footnote{\url{https://github.com/huggingface/transformers}}, i.e., \textit{BART-base-chinese}\footnote{\url{https://huggingface.co/fnlp/bart-base-chinese}} and \textit{mBART-50}\footnote{\url{https://huggingface.co/facebook/mbart-large-50-many-to-many-mmt}}.
During fine-tuning, we set the batch size to 2 and 1 for BART and mBART, respectively. All models are fine-tuned for 5 epochs with 5e-5 learning rates.
We initialize LED with the weights of BART via an official script\footnote{\url{https://github.com/allenai/longformer/blob/master/scripts/convert_bart_to_longformerencoderdecoder.py}}.
To initialize mLED with the weights of mBART, we utilize another script \footnote{\url{https://github.com/SCNUJackyChen/mBART50Long}}.

\end{sloppypar}

\subsection{Evaluation Metrics}
\begin{sloppypar}
\noindent \textbf{Automatic Evaluation.}

To comprehensively evaluate model performance, we adopt multiple automatic metrics as follows:
\begin{itemize}[leftmargin=*,topsep=0pt]
\item \textbf{ROUGE}~\cite{lin-2004-rouge}. ROUGE-N (R-N) evaluates the recall based on N-gram overlaps between the generated summaries and the corresponding references. ROUGE-L (R-L) is designed to find the length of the longest common subsequence.
\item \textbf{BLEU}~\cite{2002BLEU}. BLEU-N computes the precision based on N-gram overlaps between the generated summaries and the references.
\item \textbf{METEOR}~\cite{2005METEOR}. METEOR evaluates the harmonic mean of precision and recall, and recall weights more than precision.
\item \textbf{CIDEr}~\cite{2015CIDEr} introduces TF-IDF~\cite{jones1973index} to assign weights to n-grams and low frequency words are given higher weights than high frequency words.
\item \textbf{BertScore (B-S)}~\cite{bert-score} evaluates the semantic similarity between the generated summaries and the references.
\end{itemize}
\noindent \textbf{Human Evaluation.} For further evaluation of baselines' performance, we conduct human evaluation from three aspects:
\begin{itemize}[leftmargin=*,topsep=0pt]
\item \textbf{Coherence (Cohe.)} evaluates the quality of the generated summaries' logic and consistency.
\item \textbf{Relevance (Rel.)} evaluates the relevance between the generated summaries and the reference.
\item \textbf{Conciseness (Conci.)} evaluates how brief but comprehensive the generated summaries are.
\end{itemize}
\end{sloppypar}
\subsection{Main Results}
Table~\ref{tab:main_result} shows the experimental results. We first analyze the performance of pipeline baselines and then compare them with the end-to-end baseline.

\noindent \textbf{Pipelines.} There are two paradigms in pipeline baselines, which are extract-then-translate (Ext-Trans) and abstract-then-translate (Abs-Trans). We find that the Abs-Trans methods generally outperform the Ext-Trans methods. It is because abstractors are more flexible to generate new words or phrases based on the important sentences in documents while the extractors cannot make any modifications to the extracted sentences. Besides, the pipeline methods' performance is highly related to the adopted MT methods. 
Specifically, we equip every extractor or abstractor with Baidu and OPUS-MT MT methods, respectively. Based on the same extractor/abstractor, the performance of using the Baidu MT service is much better than that of using the OPUS-MT model.

\noindent \textbf{End-to-End vs Pipelines.} The end-to-end model achieves the best performance among all baselines. The mLED model is trained with both translation and summarization in an end-to-end manner, and thus, does not suffer from the error propagation issue.

\begin{table*}[ht]

\centering
\resizebox{0.75\textwidth}{!}
{
\begin{tabular}{lcccccccccccc}
\toprule[1pt]
\multicolumn{2}{c|}{\textbf{Method}}                 & \textbf{R-1}        & \textbf{R-2}        & \textbf{R-L}      & \textbf{B-1}  & \textbf{B-2}  & \textbf{B-3}  & \textbf{B-4} & \textbf{M} & \textbf{C} & \textbf{B-S}  

\\ \bottomrule[1pt]

\multicolumn{12}{c}{Ext-Trans}  \\ \toprule[1pt]
\multirow{2}{*}{Longest}     & \multicolumn{1}{l|}{OPUS-MT} 
& 17.5             
& 3.3             
& 14.4     
& 19.8
& 3.4
& 0.5
& 0.1
& 16.0
& 14.1
& 80.0     \\
                             & \multicolumn{1}{l|}{Baidu-MT}
& 21.6       
& 5.8     
& 18.4       
& 16.2
& 4.6
& 1.3
& 0.4
& 25.1
& 43.1
& 82.3                              \\ \hline
\multirow{2}{*}{TextRank~\cite{mihalcea-tarau-2004-textrank}}    & \multicolumn{1}{l|}{OPUS-MT} 
& 19.3    
& 3.9    
& 16.0   
& 22.5
& 4.2
& 0.7
& 0.2
& 17.2
& 19.1
& 81.0        \\
                             & \multicolumn{1}{l|}{Baidu-MT}
& 23.4                                              
& 6.7                                                
& 20.1
& 16.7
& 5.1
& 1.5
& 0.5
& 26.0
& 48.3
& 82.7          \\ \hline
\multirow{2}{*}{PacSum~\cite{zheng-lapata-2019-sentence}}      & \multicolumn{1}{l|}{OPUS-MT}  
& 18.9               
& 3.4                               
& 16.1     
& 29.2
& 4.8
& 0.7
& 0.1
& 15.8
& 18.6
& 82.1               \\
                             & \multicolumn{1}{l|}{Baidu-MT} 
& 22.3                               
& 5.3                  
& 19.1   
& 26.6
& 5.9
& 1.4
& 0.4
& 20.9
& 39.0
& 82.8        \\ \hline
\multirow{2}{*}{SummaRu.~\cite{Nallapati2017SummaRuNNerAR}} & \multicolumn{1}{l|}{OPUS-MT}  
& 19.7                              
& 4.1                           
& 17.0                            
& 30.3
& 5.3
& 0.9
& 0.2
& 16.4
& 21.5
& 82.7\\
                             & \multicolumn{1}{l|}{Baidu-MT} 
& 24.1                           
& 7.0                           
& 20.9                               
& 27.6
& 6.5
& 1.9
& 0.7
& 21.5
& 44.0
& 83.6
\\ \bottomrule[1pt]
\multicolumn{12}{c}{Abs-Trans}  \\ \toprule[1pt]
\multirow{2}{*}{PGN~\cite{see-etal-2017-get}}         & \multicolumn{1}{l|}{OPUS-MT}  
& 21.2                          
& 4.9                           
& 17.5                               
& 32.3
& 6.0 
& 1.2
& 0.3
& 17.1
& 24.5 
& 83.0\\ 
                             & \multicolumn{1}{l|}{Baidu-MT} 
& 27.6                                
& 7.9                               
& 22.9                               
& 35.1
& 8.8
& 3.1
& 1.2
& 22.5
& 55.1 
& 83.8\\ \hline
\multirow{2}{*}{LED~\cite{2020Longformer}}   & \multicolumn{1}{l|}{OPUS-MT}  
&  22.8                              
&  5.1                              
&  19.0       
&  33.7
&  6.9
&  1.5
&  0.4
&  17.4
&  27.0
&  83.5              \\ 
                             & \multicolumn{1}{l|}{Baidu-MT} 
&  29.9                              
&  9.3                              
&  25.1          
&  36.0
&  10.5
&  3.5
&  1.4
&  23.2
&  58.3
&  84.6         \\ \bottomrule[1pt]
\multicolumn{12}{c}{End-to-End} \\ \toprule[1pt]
\multicolumn{2}{c|}{mLED}      
&  \bf{32.8}                             
&  \bf{10.9}                              
&  \bf{28.7}              
&  \bf{46.7}
&  \bf{15.3}
&  \bf{5.9}
&  \bf{2.6}
&  \bf{23.7}
&  \bf{66.5}
&  \bf{85.7}       \\  \bottomrule[1pt]
\end{tabular}
}
\setlength{\belowcaptionskip}{5pt}
\caption{Experimental results (R-1/2/L: ROUGE-1/2/L; B-1/2/3/4: BLEU-1/2/3/4; M:METEOR; C:CIDEr; B-S: BertScore.).} 
\label{tab:main_result}
\end{table*}

\subsection{Generalization}
\begin{sloppypar}
To evaluate the generalization of long-document CLS models, we test the trained SummaRu., LED and mLED on the OOD test set. As the results shown in Table~\ref{tab:ood_results}, we find that models trained on Perseus do not perform as well on the OOD set, revealing the limited generalization capability of current models. 
\end{sloppypar}

\begin{table}[t]

\centering
{
\begin{tabular}{lccccccc}
\toprule[1pt]
\multicolumn{2}{c|}{\textbf{Method}}                 & \textbf{R-1}        & \textbf{R-2}        & \textbf{R-L}       & \textbf{B-S} \\ \midrule[1pt]
\multirow{2}{*}{SummaRu.} & \multicolumn{1}{l|}{OPUS-MT}
& 9.8
& 2.8
& 9.1
& 77.7\\
                        & \multicolumn{1}{l|}{Baidu-MT}
& 10.7
& \bf{3.0}
& 9.8
& \bf{80.7}\\
\midrule[1pt]

\multirow{2}{*}{LED} & \multicolumn{1}{l|}{OPUS-MT}
& 4.9
& 1.0
& 4.6
& 74.5\\
                        & \multicolumn{1}{l|}{Baidu-MT}
& \bf{12.0}
& 2.9
& \bf{11.2}
& 77.2\\
\midrule[1pt]
\multicolumn{2}{c|}{mLED}
& 7.6
& 1.4
& 7.2
& 78.5\\ \bottomrule[1pt]
\end{tabular}
}
\setlength{\belowcaptionskip}{5pt}
\caption{Out-of-domain testing results.}
\label{tab:ood_results}
\end{table}

\subsection{Human Study}
We conduct human studies under SummaRu.+Baidu, LED+Baidu and mLED correspond to extract-trans, abstract-trans, and end-to-end paradigms, respectively. We randomly select 50 samples from the in-domain test set and employ four crowd workers who are fluent in both English and Chinese to evaluate the generated summaries. The scoring adopts a 3-point scale. The final average scores are shown in Table~\ref{tab:human_study}. mLED performs better than other methods in all three aspects, indicating its strong ability to generate logical, informative and concise summaries.

\begin{table}[t]
    \centering
    \begin{tabular}{l|c|c|cc}
    \toprule[1pt]
    \multicolumn{1}{c|}{\textbf{Method}} & \textbf{Cohe.} & \textbf{Rel}. & \textbf{Conci.} \\
    \midrule[1pt]
    \multicolumn{1}{c|}{SummaRu. + Baidu-MT} & 0.82 & 1.24 & 1.12 \\

    \multicolumn{1}{c|}{LED + Baidu-MT} & 1.48 & 1.64  & 1.54 \\

    \multicolumn{1}{c|}{mLED} & \bf{1.62} & \bf{1.72} & \bf{1.58}\\
    \bottomrule[1pt]
        
    \end{tabular}
    \setlength{\belowcaptionskip}{5pt}
    \caption{The results of human study on Perseus.}
    \label{tab:human_study}
\end{table}

\begin{figure*}[ht]
    \centering
    \includegraphics[width=17cm]{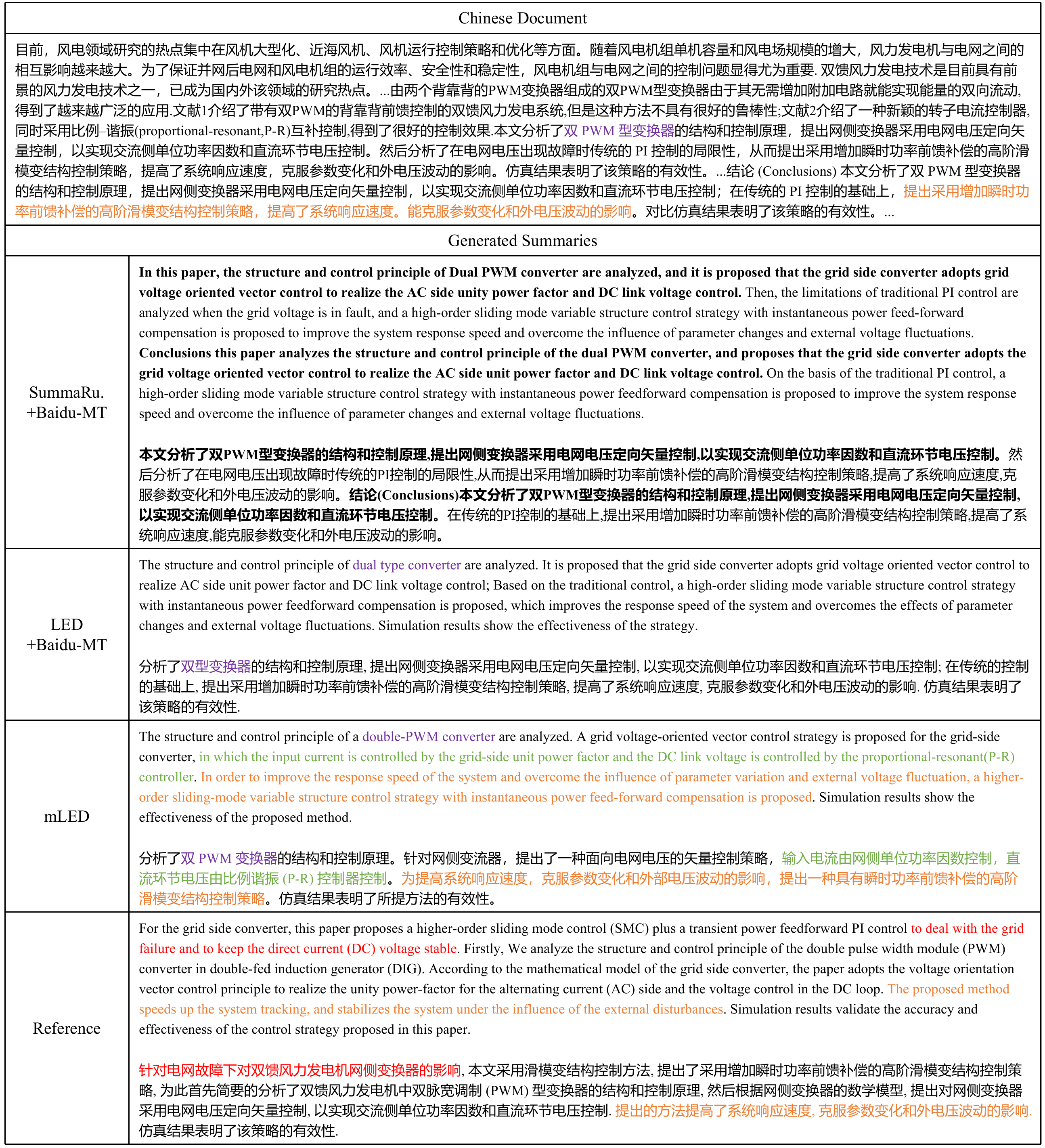}
    \caption{An example of Perseus and the generated summary of strong baselines.}
    \label{fig:case_study}
\end{figure*}
\subsection{Case Study}
We give an example from our dataset and show the generated summaries of several strong baselines. The example is shown in Figure~\ref{fig:case_study}.
For pipeline methods, the Chinese summaries are directly generated by monolingual summarization models, and the English summaries are translated from the corresponding Chinese summaries.
The end-to-end model (mLED) directly generates the English summary, and the Chinese summary is translated from the English one.
For SummaRu., the source summary is obtained by concatenating several key sentences that are directly extracted from the source document.
In this case, we find that the two bold sentences are the same, although they are from different parts of the document, revealing the problem that extractive approaches could cause semantic duplication.
For LED, it generates Chinese summaries from Chinese documents. Due to its monolingual characteristics, the presence of content in other languages in the source documents may result in generation errors. As the purple parts showed in Figure~\ref{fig:case_study}, LED misses the English part of the professional term, leading to semantic errors, while mLED performs well in this aspect because of its strong ability to deal with the cross-lingual setting.
For mLED, it generates error messages that do not conform to the source document. For example, the green sentence is not contained in the source document and there is no content in the source document expressing the meaning of this sentence. Moreover, the orange sentence in LED's summary means \textit{improving the response speed of the system and overcome the influence of parameter variation and external voltage fluctuation} is the purpose of \textit{proposing a higher-order sliding-mode variable structure control strategy with instantaneous power feed-forward compensation}, while in both of the source document and the reference summary, the former is a consequence of the latter. 
The red part in the reference summary does not appear directly in the source document, and it needs to be inferred from the source document. However, both LED and mLED miss the information, indicating their limited understanding and reasoning abilities.

\section{Discussion}
\begin{sloppypar}

To further understand the challenges of our long-document CLS dataset, we randomly select 200 samples from the test set and analyze the wrong generation of mLED.
We list four main errors that occur in the generated summaries as follows:
\begin{itemize}[leftmargin=*,topsep=0pt]
\item \textbf{Missing information}: the generated summaries neglect some information involved in the references.
\item \textbf{Redundancy}: the generated summaries have additional information that does not exist in the references.
\item \textbf{Wrong references}: some information in generated summaries is not faithful to the source documents.
\item \textbf{Semantically unclear generation}: the generated summaries contain information that is incomprehensible.
\item \textbf{Other}: the errors do not belong to any of the former error types.
\end{itemize}

\begin{table}[ht]
    \centering
    \begin{tabular}{l|c}
    \toprule[1pt]
         \textbf{Error} & \textbf{\%} \\ \midrule[1pt]
         Missing information     &   87.5  \\
         Redundancy              &   40.5  \\
         Wrong references        &   24.5  \\
         Semantically unclear generation    &   22.0  \\
         Other                   & 17.5 \\
    \bottomrule[1pt]
    \end{tabular}
    \setlength{\belowcaptionskip}{5pt}
    \caption{Main error types of mLED's outputs. Note that each generated summary might contain multiple errors.}
    \label{tab:error_analysis}
\end{table}

Table~\ref{tab:error_analysis} shows the proportion of each error type. We find that (1) the proportions of missing information and redundancy are higher than others, indicating that it is difficult for mLED to grasp key information from long documents. (2) The wrong references problem also occupies a certain proportion. Being faithful to the source documents is very important in the summarization task, especially in the field of scientific papers, but mLED still has some problems in this regard.
(3) Semantically unclear generation problem shows that mLED has insufficient ability to generate long sequences.

\vspace{0.5ex}
\noindent \textbf{Missing information} and \textbf{Redundancy}. This is mainly caused by the long sequences of the input documents. Besides, the compression ratio of CLS samples in Perseus is also at a high level, which also means the information in the summaries is sparsely distributed across documents. There, it is non-trivial for models to generate informative and accurate summaries for documents.
In addition, although mLED can process long sequences, its attention mechanism is essentially local attention (windowed local attention with global attention), which has a limited capacity of interacting between long distant content.
As a result, mLED cannot fully incorporate information from the entire document and extract the truly important information.
In severe cases, it will extract a series of content that is internally correlated but irrelevant to the ground truth summary.
To alleviate the problems of missing information and redundancy, the model needs to integrate the information of the entire document and distinguish between core and non-core information.
Future work could integrate the information of documents by introducing hierarchical structures, e.g., dividing and linking different parts of documents. 

\vspace{0.5ex}
\noindent \textbf{Wrong references}. mLED generates information that does not conform to the references. For example, a reference summary says ``Methods: Fifty-eight patients with subaortic stenosis were treated surgically in our center from December 1996 to October 2019.'', but the generated summary is ``Methods: The clinical data of 13 patients with congenital heart disease were retrospectively analyzed.''.
The wrong reference is mainly caused by the long-distance dependencies problem. When generating summaries, the model needs to fuse information across long distances. However, as the distance grows, the long-distance information becomes more and more blurred, leading to generating wrong information.

\vspace{0.5ex}
\noindent \textbf{Semantically unclear generation}. mLED generates sentences like ``data mining is an important content of data mining.'' and ``the results provide us with a basis to judge whether the sub-time series of time series with increasing and decreasing is the sub-time series with great increasing and decreasing.'' that involve wrong syntax and are difficult to understand.
The long-distance dependencies problem is one of the reasons for this error. In addition, this error also exposes the inadequacy of the current generative models in generating long texts.

\vspace{0.5ex}
For wrong references and semantically unclear generation, it is important to address the problem of long-distance dependencies. The best way is to create a pre-trained language model that can efficiently incorporate long document information, but it is non-trivial to do so~\cite{shi-etal-2022-layerconnect,10.1007/978-3-031-00129-1_6}.
Instead, we think it is possible to transform long-document CLS task into short-document CLS task by combining extractive methods with abstractive methods.
For example, an extractive method is used to extract sufficient key sentences, and then a well-performed multilingual seq2seq model, such as mBART, is adopted to generate a summary based on the extracted sentences.
Although it is a pipeline method and there is error propagation problem, but it can avoid the problem of long-distance dependencies. Moreover, the first step of this method can filter out irrelevant information to a large extent, and the self-attention of mBART can be fully utilized to fuse the entire content.

To summarize, our long-document CLS dataset, Perseus, brings a lot of new challenges to the CLS task: (1) its documents are too long for current CLS models to process; (2) the summaries are also relatively long and sparsely distributed across documents, making it difficult to generate; (3) the length of the document makes it difficult for the model to incorporate the entire document information; (4) the problem of long-distance dependencies makes it hard for seq2seq models to generate correct summaries; (5) it is hard to process the professional terms correctly.

\end{sloppypar}

\vspace{0.5ex}

\section{Conclusion}
\begin{sloppypar}
In this paper, we introduce the long-document cross-lingual summarization (long-document CLS) task and propose the first long-document CLS dataset, Perseus. We conduct multiple experiments on our dataset and analyze the advantages and disadvantages of different summarization methods.
To evaluate the generalization of long-document CLS models trained on our dataset, we also provide an out-of-domain test set which is in the sports domain.
To further understand the challenges brought by Perseus, we manually analyze the generated summaries of mLED, take a deep dive into the reasons behind these errors and discuss the possible solutions. 
In the future, we would like to focus on expanding the multilingual version of Perseus to meet the needs of different languages and explore a more efficient method for long-document CLS tasks.
\end{sloppypar}

\section*{Acknowledgements}
We would like to thank anonymous reviewers for their suggestions and comments. This work is supported by the National Natural Science Foundation of China (No.62072323, 62102276), Shanghai Science and Technology Innovation Action Plan (No. 22511104700), the Natural Science Foundation of Jiangsu Province (Grant No. BK20210705), and the Natural Science Foundation of Educational Commission of Jiangsu Province, China (Grant No.
21KJD520005).

\vfill
\newpage

\bibliographystyle{ACM-Reference-Format}
\bibliography{references}

\appendix

\end{document}